\documentclass[conference]{IEEEtran}

\usepackage{times}
\usepackage[numbers]{natbib}
\usepackage{multicol}
\usepackage[bookmarks=true]{hyperref}
\usepackage{xcolor}
\usepackage{xspace}
\usepackage{amsmath}
\usepackage{amssymb}
\usepackage{booktabs}
\usepackage{multirow}
\usepackage{bbm}
\usepackage[algo2e,algoruled,boxed,lined,noend]{algorithm2e}
\usepackage{wrapfig}
\usepackage{float}
\usepackage{graphicx}
\usepackage{subcaption}
\usepackage{hyperref}
\hypersetup{
    colorlinks=true,
    linkcolor=orange,
    filecolor=magenta,      
    urlcolor=orange,
    citecolor=orange,
}
\usepackage[capitalise, nameinlink]{cleveref}

\newcommand{\myfigref}[1]{Figure~\ref{#1}}

\newcommand{\myeqref}[1]{Equation~(\ref{#1})}
\newcommand{\mysecref}[1]{Section~\ref{#1}}
\newcommand{\mytabref}[1]{Table~\ref{#1}}

\definecolor{darkgreen}{rgb}{0.09, 0.45, 0.27}
\newcommand{\bestscore}[1]{\textcolor{darkgreen}{\mathbf{#1}}}

\NewDocumentCommand\emojishout{}{
    \includegraphics[scale=0.08]{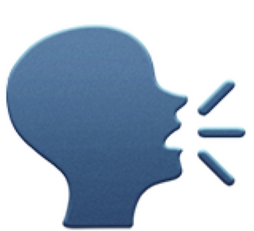}
}

\newcommand{\ours}[0]{YAY Robot\xspace}
\newcommand{\full}[0]{Yell At Your Robot\xspace}
\newcommand{\piL}[0]{\pi_{\text{L}}}
\newcommand{\piH}[0]{\pi_{\text{H}}}

\begin{document}

\title{\full\emojishout\\Improving On-the-Fly from Language Corrections}

\author{\authorblockN{Lucy Xiaoyang Shi$^1$\quad Zheyuan Hu$^2$\quad Tony Z. Zhao$^1$\quad Archit Sharma$^1$\quad \\Karl Pertsch$^{1,2}$\quad Jianlan Luo$^2$\quad Sergey Levine$^2$\quad Chelsea Finn$^1$}
\authorblockA{$^1$Stanford University\qquad$^2$UC Berkeley}
}

\makeatletter
\let\@oldmaketitle\@maketitle%
\renewcommand{\@maketitle}{\@oldmaketitle%
  \begin{center}
  \captionsetup{type=figure}
  \includegraphics[width=\textwidth]{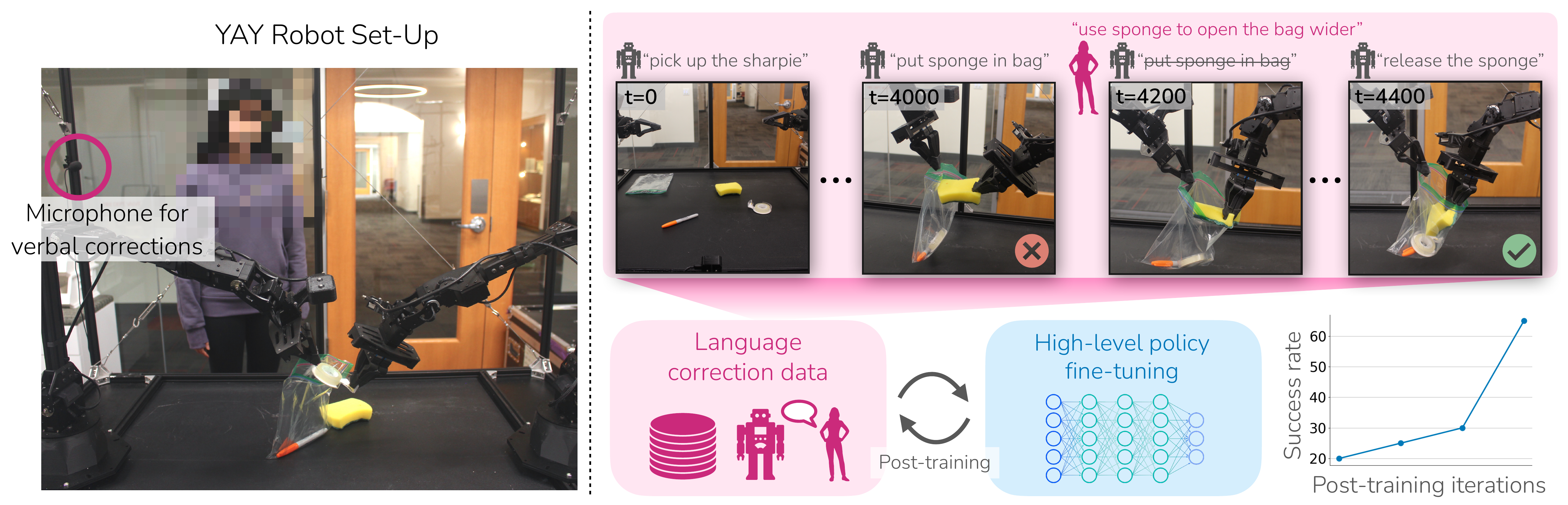}
    \captionof{figure}{Our approach enables robots to leverage verbal corrections to improve performance on complex long-horizon tasks like packing a ziploc bag and preparing trail-mix. It can incorporate verbal corrections in real-time (top) and for continuous improvement (bottom). } 
    \vspace{-1.0em}
    \label{fig:teaser}
    \end{center}
}
\makeatother

\maketitle

\begin{abstract}
Hierarchical policies that combine language and low-level control have been shown to perform impressively long-horizon robotic tasks, by leveraging either zero-shot high-level planners like pretrained language and vision-language models (LLMs/VLMs) or models trained on annotated robotic demonstrations. However, for complex and dexterous skills, attaining high success rates on long-horizon tasks still represents a major challenge -- the longer the task is, the more likely it is that some stage will fail. 
In principle, a robust high-level controller can compensate for low-level failures by dynamically deploying corrections and adjustments, but training such high-level controllers in a way that is aware of the physical capabilities of the low-level skills requires costly demonstrations of entire multi-stage tasks. 
Can humans help the robot to continuously improve its long-horizon task performance through intuitive and natural feedback?
In this paper, we make the following observation: high-level policies that index into sufficiently rich and expressive low-level language-conditioned skills can be readily supervised with human feedback in the form of language corrections. 
We show that even fine-grained corrections, such as small movements (``move a bit to the left''), can be effectively incorporated into high level policies, and that such corrections can be readily obtained from humans observing the robot and making occasional suggestions.
This framework enables robots not only to rapidly adapt to real-time language feedback, but also incorporate this feedback into an iterative training scheme that improves the high-level policy's ability to correct errors in both low-level execution and high-level decision-making purely from verbal feedback.
Our evaluation on real hardware shows that this leads to significant performance improvement in long-horizon, dexterous manipulation tasks without the need for any additional teleoperation. 
Videos and code are available at \href{https://yay-robot.github.io/}{https://yay-robot.github.io/}.
\end{abstract}

\IEEEpeerreviewmaketitle

\section{Introduction}
\label{sec:introduction}

Complex robotic tasks may require sequencing multiple individual primitives. For example, packing multiple items into a bag, as shown in~\myfigref{fig:teaser}, requires grasping each object in turn, maneuvering it near the bag opening, and inserting it. An appealing framework for addressing such multi-stage tasks is via \textit{hierarchical abstraction}, where a high-level policy commands specific behaviors that are then performed by the low-level policy~\citep{luo2023multistage, xie2020deep, yu2021multi, duan2019learning}.  
One intuitive method for parameterizing such policies is via language, with a high-level policy selecting
among possible language instructions at each stage~\citep{brohan2022rt1, ahn2022say}. Unfortunately, as the number of stages in a task increases, there are more points of failure -- if every stage needs to succeed, the overall probability of failure goes up exponentially. However, a robust high-level policy can compensate for low-level failures, deploying corrections and adjustments as needed. Therefore, successful completion of such multi-stage tasks depends critically on the ability to train such high-level policies in a way that is robust, aware of the limitations of the low-level primitives, and well adapted to the dynamics of the problem. 

Unfortunately, this is difficult to do in a scalable way. Complete end-to-end on-robot demonstrations of long-horizon tasks provide ``gold standard'' supervision, because they allow the high level to be fully aware of the intricacies of low-level control, but they are costly and time consuming to collect at scale because the tasks are so long. Knowledge transfer from other sources, such as large language models (e.g., as in works that use LLMs or VLMs for planning~\citep{ahn2022say, huang2023voxposer, huang2022language}) provides an appealing alternative, but this knowledge is not grounded in robotic behaviors, leaving the high-level policy relatively brittle because it does not know which skills are more or less effective \emph{for this particular robot and in this particular situation}. Other modes of indirect supervision, such as language-only supervision~\citep{driess2023palme} or human videos~\citep{mendonca2023structured, bahl2023affordances, sivakumar2022robotic} similarly provide indirect supervision. Therefore, the challenge of training robust high-level policies can be seen as the challenge of obtaining scalable and high-quality training data for the high level.

What if instead of requiring extensive demonstration supervision for such high-level policies, we could instead train them with natural feedback from humans in the form of language corrections? Such feedback is natural for humans to provide, and might even be gathered naturally in the course of the robot's day-to-day work, as it attempts the desired tasks and receives feedback from human users. Our key insight in this work is that this sort of language feedback can be readily incorporated into hierarchical policies when the high-level policy outputs language commands, and in fact this procedure can be seen as a high-level analogue of the widely used DAgger~\citep{ross2010reduction, kelly2018hgdagger} algorithm -- a technique where the robot iteratively incorporates feedback from a human supervisor, typically in the form of low-level motor actions -- but over the action space of a high-level instruction policy, i.e., over language instructions.
By incorporating language \emph{corrections}, the high-level policy is equipped to correct mistakes made by both the low-level policy and the high-level policy. 
This facilitates robust recovery from failures, which is critical for the success in long-horizon tasks.

Based on this insight, we propose \full (\ours), a system for improving robot post-training through natural language feedback. 
Unlike previous efforts that focus on post-hoc action relabeling or one-time correction~\citep{liu2023interactive, cui2023right}, \ours aims for a more organic integration of language into policy improvement through: (a) on-the-fly adaptation to language feedback, and (b) continuous improvement from user interaction.

First, we enable robots to adapt their behaviors to diverse, often contextual language commands in real time.
We use an end-to-end Language-Conditioned Behavior Cloning (LCBC) policy to learn a variety of skills specified through language within a single neural network, facilitating versatile low-level behaviors in response to diverse language inputs.
As a result, users can interact with our system using free-form language, ranging from low-level commands like ``move the sponge a little lower'' or ``turn your right gripper towards me'' to high-level preferences such as ``I want more M\&M’s in my trail mix bag''.

Moreover, robots should improve from user feedback over time, avoiding perpetual corrections and learning to handle complex, long-horizon tasks more effectively without frequent interventions.
Hence, we learn a high-level policy to mimic human instruction patterns and assimilate human feedback. 
As illustrated in~\myfigref{fig:teaser}, our high-level policy generates language instructions for the low-level LCBC policy autonomously. 
When users choose to intervene, their commands temporarily override the high-level policy and directly feed into the low-level policy, allowing for immediate adaptation from the robot. 
After user interactions, the high-level policy is finetuned on human language interventions to enhance its ability to predict better instructions and corrections in the future.

The primary contribution of our work is \ours, a learning-based system for long-horizon tasks that allows robots to both (a) incorporate language corrections on-the-fly and (b) continuously improve from this feedback. 
In our experiments, we consider three bi-manual multi-stage manipulation tasks: packing three items into a ziploc bag, making a bag of trail mix by scooping ingredients, and cleaning gummies off a plate. By leveraging human corrections on-the-fly, our approach leads to real-time improvement from 15\% to 50\%. Moreover, incorporating these corrections into high-level policy training increases performance from 15\% to 45\%. 
Crucially, \ours allows users to guide robots naturally and intuitively, representing a step towards enabling end-users to directly teach robots through natural language in everyday scenarios.

\begin{figure*}[t]
    \centering
    \includegraphics[width=.9\textwidth]{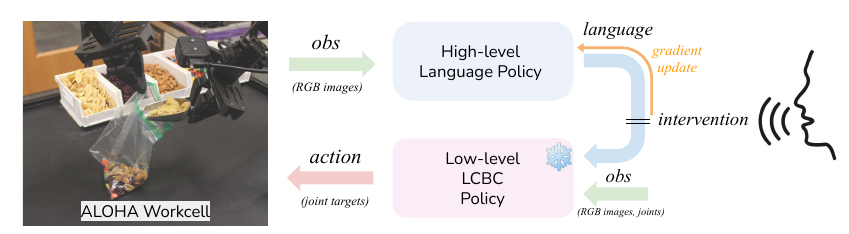}
    \caption{\textbf{Overview of \ours.} We operate in a hierarchical setup where a high-level policy generates language instructions for a low-level policy that executes the corresponding skills. During deployment, humans can intervene through corrective language commands, temporarily overriding the high-level policy and directly influencing the low-level policy for on-the-fly adaptation. These interventions are then used to finetune the high-level policy, improving its future performance.}
    \label{fig:method}
    \vspace{-5mm}
\end{figure*}
\section{Related Work}
\label{sec:related_work}
The idea of learning a single policy to complete several tasks has a rich history in robotics~\citep{da2012learning, kober2012reinforcement, deisenroth2014multi, parisotto2015actor, teh2017distral, arora2018multi, mao2023learning}. As the number of behaviors increases, natural language provides a simple and interpretable abstraction to index these behaviors, and thus, several robotic systems map natural language instructions to actions~\citep{macmahon2006walk, kollar2010toward, mei2016listen, misra2016tell, stepputtis2020language, lynch2020grounding, lynch2020language, shridhar2022cliport}. 
This idea has been explored in a rich body of work spanning semantic parsing for robotics~\citep{Kollar2010GroundingVO, tellex2011understanding, walker19neural} and task and motion planning (TAMP)~\citep{503568, GarrettCHKSKL21, 10342169}.
More recently, the advent of large robotic interaction datasets has enabled learning language-conditioned multi-task policies~\citep{jang2022bc, brohan2022rt1, brohan2023rt} that can generalize to novel scenes, objects and even natural language instructions.

Representing the low-level behaviors using natural language has several desirable properties for high-level decision making. First, natural language offers a compact and compositional representation, that can allow combinatorial generalization~\citep{shu2017hierarchical, andreas2017learning, jiang2019language}. This makes it a compelling abstraction to tackle long-horizon tasks~\citep{jiang2019language, sharma2021skill, garg2022lisa, ahuja2023hierarchical}, especially with the increasing use of large language models for high-level planning~\citep{ahn2022say, huang2022language, liang2022code, huang2023voxposer, vemprala2023chatgpt}. 
Second, language provides a natural interface for humans to guide, correct, and improve a robot by simply \textit{talking} to it. 
Human-in-the-loop imitation learning techniques often require people to intervene and correct robot behaviors using teleoperation or kinesthetic teaching~\citep{ross2010reduction, kelly2018hgdagger, mandlekar2020human, hoque2021lazydagger, hoque2021thriftydagger, liu2022robot}. Natural language is more accessible in contrast, and recent works have shown that it can enable humans to interact and guide robots in real-time~\citep{lynch2023interactive}, provide on-the fly corrections for both high-level plans~\citep{broad2017real, sharma2022correcting} or low-level behaviors~\citep{bucker2023latte, cui2023right} in a shared autonomy setup, modify the code~\citep{liang2024learning}, or even update low-level behaviors post-hoc by converting language corrections into new target actions~\citep{liu2023interactive}. Our work combines the benefits of both these properties;
\ours learns a language-conditioned low-level policy and trains a high-level policy to output natural language instructions to combine low-level skills for long-horizon tasks. Further, \ours can fine-tune the high-level policy after deployment with verbal corrections, leading to consistent improvements in performance on hard long-horizon tasks.
In contrast to OLAF~\citep{liu2023interactive}, \ours can modify robot behaviors on-the-fly from human language interventions and can learn end-to-end from raw pixel observations without any explicit state estimation. And in contrast to other works~\citep{broad2017real, bucker2023latte, cui2023right, wang2024mosaic} that are designed for shared autonomy, and thus require humans to provide corrections perpetually, \ours is designed to operate autonomously while still being able to improve from verbal corrections when provided.
One concurrent work, RT-H~\citep{belkhale2024rth}, shares a similar idea of learning from language corrections. However, RT-H’s correction is limited to a fixed set of spatial movements, whereas ours allows flexible, diverse user input such as ``use the sponge to open the bag wider'' to tackle long-horizon, bi-manual dexterous manipulation tasks.

\section{\full}
\label{sec:method}

\begin{figure*}[t]
    \centering
    \includegraphics[width=\textwidth]{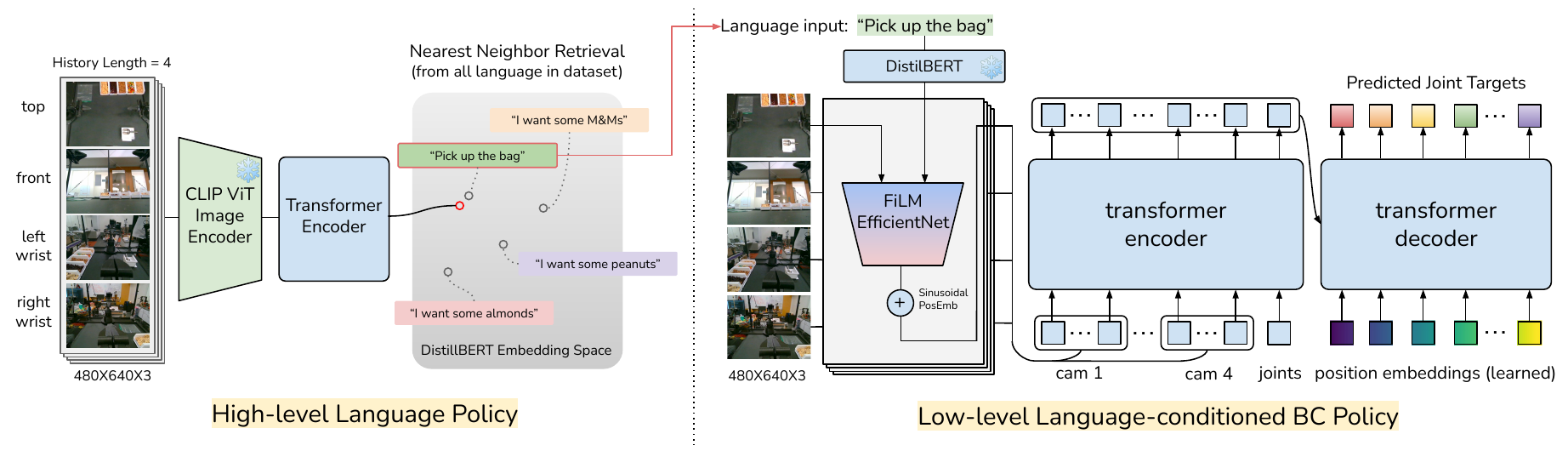}
    \vspace{-6mm}
    \caption{\textbf{Policy Architecture.} Our system processes RGB images and the robot's current joint positions as inputs, outputting target joint positions for motor actions.
    The high-level policy uses a Vision Transformer to encode visual inputs and predicts language embeddings. The low-level policy uses ACT, a Transformer-based model to generate precise motor actions for the robot, guided by language instructions. This architecture enables the robot to interpret commands like ``Pick up the bag'' and translate them into targeted joint movements.}
    \label{fig:architecture}
    \vspace{-5mm}
\end{figure*}

In this section, we describe each component of our system that enables on-the-fly and continual improvement from language corrections. We will first describe the overall problem set-up and define notation before detailing the low-level and high-level policy components. Last, but most importantly, we will describe how we integrate verbal corrections into the framework.

\subsection{Preliminaries}
\label{sec:prelim}

We formulate our robot manipulation task as a Markov Decision Process (MDP), denoted as \(\mathcal{M} = (\mathcal{S}, \mathcal{A}, \mathcal{P}, p_{0})\), where \(\mathcal{S}\) represents the state space, \(\mathcal{A}\) the action space, \(\mathcal{P}\) the transition probabilities, and \(p_{0}\) the initial state distribution. 
The robot does not have access to the true states \(s \in \mathcal{S}\); instead, it receives observations \(\mathcal{O}\) that are partially observed.

For training the robot's high-level policy \(\piH\) and low-level policy \(\piL\), we utilize a base dataset \(\mathcal{D}\), composed of sequences \((o_t, a_t, l_t)\), where \(o_t\) represents the observation, including RGB images and robotic proprioceptive information, \(a_t\) is the robot's action, and \(l_t\) is a language instruction given at time \(t\). 
This dataset is assumed to cover a broad range of visuomotor skills necessary for completing tasks, alongside various mistakes and recovery behaviors.

After the initial training phase, we finetune the robot's high-level policy to align it with human verbal feedback. 
This is done using a correction dataset \(\mathcal{D}_{\text{corr}}\) consisting of online user interaction data, \((o_t, l^{\text{user}})\) -- data that is naturally produced in the course of humans interacting with robots.
Unlike the base dataset, \(\mathcal{D}_{\text{corr}}\) omits \(a_t\) because we keep the low-level policy frozen during this post-training phase and only update the parameters of the high-level policy.

\subsection{Low-Level Language-Conditioned Behavior Cloning}
\label{subsec:ll_policy}

The low-level policy enables the robot to interpret and execute a wide array of skills articulated through natural language commands. 
Learning an expanded set of skills that are not just the minimal set for completing the task provides more flexibility, as it allows the high-level policy to have the latitude to orchestrate skills that correct for previous mistakes, and the low-level policy to accommodate these corrections and adjustments on-the-fly. 
Motivated by the need for such flexibility, the low-level policy is implemented as a deep neural network trained end-to-end on datasets encompassing diverse visuomotor skills, ranging from task-centric instructions, such as ``pour into the bag'', to task-agnostic corrections like ``move the left arm towards me''.

The low-level policy, \(\piL(a_t|o_t, l_t)\), maps the current observation \(o_t\) and language instruction \(l_t\) to action \(a_t\). 
The policy is trained using the standard Behavior Cloning (BC) objective:
\begin{equation}
\min_{\piL} \mathbb{E}_{(o_t, l_t, a_t) \sim \mathcal{D}} \left[ \mathcal{L}_{\text{BC}}(\piL(a_t|o_t, l_t), a_t) \right]
\end{equation}
where \(\mathcal{L}_{\text{BC}}\) is a loss function (e.g., \(\ell_1\) or \(\ell_2\)) comparing the predicted continuous action to the ground truth action.

\subsection{High-Level Policy for Autonomous Instruction Generation}
\label{subsec:hl_policy}

A hierarchical setup could allow the robot to reuse primitive skills.
Therefore, we learn a high-level policy to generate language instructions that guide the low-level policy. 
This policy is also trained using behavior cloning to predict language instructions conditioned on the current observation.
Similar to a vision-language model (VLM), it processes image observations and outputs language embeddings.
Since the same observation can lead to different commands (e.g., ``tilt the scoop down'' and ``go a bit higher'' are both reasonable when the robot is close to food containers), we contextualize the instructions by conditioning on a brief history of observations.

The high-level policy is denoted as \(\piH(l_t | o_{t-k:t})\), 
where \(l_t\) represents the language embedding generated at time \(t\), and \(o_{t-k:t}\) is the temporal context (i.e., a sequence of observations) up to time \(t\). 
This model is also trained end-to-end, using a cross-entropy loss to align the generated language instructions with those in the training dataset. 
The logits are obtained using cosine similarity.
The training objective can be formulated as:
\begin{equation}
\label{eq:hl}
\min_{\piH} \mathbb{E}_{(o_{t-k:t}, l_t) \sim \mathcal{D}} \left[ \mathcal{L}_{\text{CE}}(\piH(l_t | o_{t-k:t}), l_t) \right]
\end{equation}
where \(\mathcal{L}_{\text{CE}}\) is the cross-entropy loss function.

\subsection{Policy Integration and Adaptation to Human Feedback}
\label{subsec:integration_adaptation}

\ours integrates these policies to create a cohesive system capable of adapting to real-time language feedback. 
As shown in~\myfigref{fig:method}, the high-level policy generates language instructions for the low-level policy, which executes the corresponding skills. 
During deployment, the human may intervene to correct erroneous behaviors or indicate preferences by providing a corrective language command.
The user's verbal interventions temporarily override the high-level policy's output, directly influencing the low-level policy to enable on-the-fly adaptation:
\begin{equation}
    \pi_{\text{deploy}}(a_t | o_{t-k:t}) = \begin{cases} 
      \piL(a_t | o_t, l^{\text{user}}) & \text{if intervention} \\
      \piL(a_t | o_t, l^\text{H}) & \text{otherwise} 
   \end{cases}
\end{equation}
where \(l^{\text{user}}\) denotes the language command provided by the user, and \(l^\text{H} = \arg\max \piH(\cdot \mid o_{t-k:t})\).
This intervention is recorded as a new data point \((o_{t-k:t}, l^{\text{user}})\) in a correction dataset \(\mathcal{D}_{\text{corr}}\), which is subsequently used to finetune \(\piH\).

\subsection{Continuous Improvement from Human Feedback}
\label{subsec:continuous_improvement}

Building upon the real-time adaptation capability, \ours aims to learn continuously to minimize the need for perpetual corrections and better align with user preferences over time. 
The continuous improvement of \ours is driven by incorporating human language feedback into the high-level policy. 
This process is conceptually similar to performing Human-Gated DAgger (HG-DAgger~\citep{kelly2018hgdagger}) on the high-level policy, whose actions are language commands. 
Crucially, unlike typical uses of DAgger and HG-DAgger, interventions are only provided in \textit{natural language} rather than low-level robot actions.
The low-level policy is kept frozen throughout this post-training process.

\subsubsection{Finetuning High-Level Policy}
We finetune the high-level policy on the correction dataset \(\mathcal{D}_{\text{corr}}\) along with the base dataset \(\mathcal{D}\). 
The combined dataset \(\mathcal{D} \cup \mathcal{D}_{\text{corr}}\) enhances the policy's exposure to diverse scenarios that result from mistakes by either the low-level or high-level policy, both of which necessitate corrections. 
This process aligns the policy's predictions with both the initial training instructions and the human corrective feedback.
The optimization objective is the same as in \myeqref{eq:hl}.

\subsubsection{Iterative Improvement}
After each iteration of user interaction and feedback collection in \mysecref{subsec:integration_adaptation}, we fine-tune the policy to reflect the new data. This iterative process can be depicted as:
\begin{equation}
\piH^{(n+1)} = \text{Post-Training}(\piH^{(n)}, \mathcal{D} \cup \bigcup_{i=1}^n  \mathcal{D}_{\text{corr}}^{(i)})
\end{equation}
Here, \(\piH^{(n)}\) and \(\piH^{(n+1)}\) denote the high-level policy before and after the \(n\)-th iteration of finetuning, respectively. \(\mathcal{D}_{\text{corr}}^{(n)}\) is the dataset of corrective feedback obtained at the \(n\)-th iteration. 
This fine-tuning process ensures that \ours progressively improves in handling complex tasks autonomously without frequent interventions.

\section{Practical Instantiation and Implementation}
\label{sec:implementation}

This section outlines our implementation, focusing on data collection, policy architecture, and the post-training procedure.

\subsection{Pre-Training Data Collection and Processing}
\label{subsec:data_collection}

\subsubsection{Language Annotation}
We need language-annotated robotic data to pre-train the base policy. 
Traditional methods involve post-hoc language annotation, where operators watch robot videos and annotates each skill segment with start and end timestamps. 
However, this process is laborious, particularly for long-horizon tasks involving numerous skill segments.

To streamline this process, we adopt a more efficient data collection method: live narration. 
By placing a microphone near the robot, the operator can narrate the skill they are performing in real-time. 
They first verbalize the intended skill and then teleoperate the robot to perform it. 
The recorded audio is then transcribed into text using the Whisper~\citep{radford2022robust} model and synchronized with the robot's trajectory.
Our code, which automates this process, is open-sourced to facilitate research that involves language-annotated data collection.

\subsubsection{Filtering Mistakes}
Our data includes both successful executions and errors which lead to subsequent recoveries. 
If trained on all the data, a model might learn to intentionally make mistakes (\mysecref{subsec:ablation}). An intuitive idea is to filter out the segments preceding corrections, since they are likely erroneous or suboptimal and should be excluded from training.
However, identifying errors traditionally requires expensive post-hoc labeling. 
We simplify this by differentiating instructions from corrections. 

To implement this distinction, we use foot pedals during data collection. 
When narrating a new skill (instruction), the operator steps on the instruction pedal. 
If a correction is necessary, they step on the correction pedal. 
This method allows for quick filtering of segments leading up to corrections, ensuring they are not used for training. 

\subsubsection{Collection of Low-Level Correction Skills}
For dexterous, long-horizon tasks, operators naturally make mistakes, and their recovery behavior informs the collection of correction skills. 
Nonetheless, the situations where the operator makes mistakes and the robot makes mistakes could be different. 
Therefore, to ensure the collection of relevant and practical correction skills for robots, we assess policy performance to identify which skills to collect. 
We train robots using existing data, and during policy rollouts, whenever an operator observes, for example, ``I wish I could just tell the robot to adjust its gripper slightly'' when the robot behaves suboptimally, such corrections are narrated and included in future training data collection.

\subsection{Low-Level Policy}
\label{subsec:ll_policy_implementation}

The low-level policy is designed to perform precise and complex robotic actions based on both visual and language inputs. 
To achieve this, we use Action Chunking with Transformers (ACT)~\citep{zhao2023learning}, a state-of-the-art imitation learning architecture that has demonstrated efficacy in handling robotic tasks requiring high precision and robustness~\citep{zhao2023learning}.
For visual processing, we use EfficientNet~b3~\citep{tan2019efficientnet} as the visual backbone to encode RGB images captured from each camera. 
In the original ACT model, image features are concatenated with the robot's proprioceptive state information as the input to the policy. 
To additionally condition on language, we modify ACT by incorporating FiLM~\citep{perez2017film} layers to fuse the visual and language inputs, similar to RT-1~\citep{brohan2022rt1}. 
We encode language instructions in the dataset using DistilBERT~\citep{sanh2019distilbert}.

\subsection{High-Level Policy}
\label{subsec:hl_policy_implementation}

The high-level policy generates language commands autonomously. 
This policy is based on a visual backbone using a Vision Transformer (ViT)~\citep{dosovitskiy2020image} initialized with pretrained CLIP~\citep{radford2021learning} weights. 
These weights are kept frozen throughout training. 
The visual feature is then processed through additional Transformer~\citep{vaswani2017attention} and MLP layers to produce a language encoding. 
For ground truth encoding, we utilize DistilBERT~\citep{sanh2019distilbert} to process language instructions in the dataset.

To account for temporal context, the model employs sinusoidal position embeddings. 
These are applied to a sequence of historical observations for each camera at every timestep. 
In selecting history observations, in case of high-frequency control,  the immediate previous image has high similarity to the current frame. 
Instead of the immediate previous images, we choose up to four images spaced one second apart, providing a broader temporal context.

During training, the output language embedding is compared to existing language commands in the dataset using cosine similarity. This comparison yields logits, which are then evaluated using cross-entropy loss with a learned temperature parameter~\citep{oreshkin2018tadam}. 
To encourage the model to predict upcoming instructions rather than current ones, we offset the target prediction by a small margin to increase the likelihood of the model predicting the next instruction as it nears the end of the current skill segment.
At inference time, the most probable language command is selected as the high-level policy output.

\begin{figure*}[t]
    \centering
    \includegraphics[width=\textwidth, height=0.75\textheight, keepaspectratio]{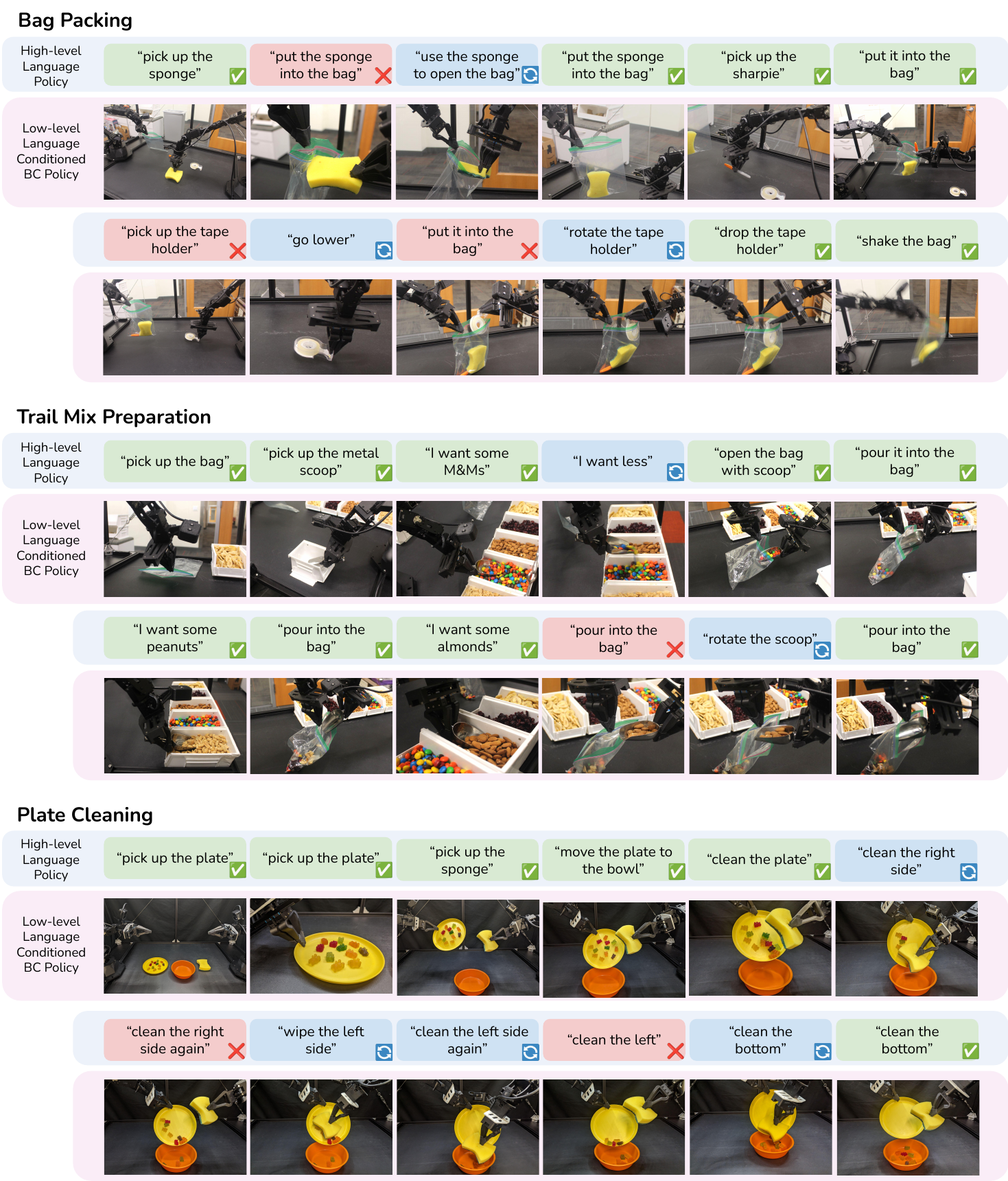}
    \vspace{-0.25em}
    \caption{\textbf{Real-World Task Rollouts with Language Corrections.} For each of the 3 long-horizon bimanual manipulation tasks, we illustrate the sub-tasks, common failure modes, examples of verbal corrections, and robot's corrective behaviors.}
    \label{fig:task}
    \vspace{-1em}
\end{figure*}

\subsection{Post-Training}
\label{subsec:post_training}

During post-training, the high-level policy is queried at fixed intervals, specifically every 4 seconds as the average skill length, to generate language instructions for the robot.

To facilitate real-time human feedback, a microphone is placed near the robot to capture verbal commands from users.
If the user wishes to intervene, they can verbally instruct the robot to stop with a simple command such as ``stop'' (or a more polite alternative like ``pardon'').
Following this interruption, the user can provide verbal correction to guide the robot.

To continually improve the high-level policy, we record the verbal corrections provided by the user along with the corresponding observations. 
Additionally, considering the human reaction time, the system also saves data from 2 seconds prior to the intervention for more context.
This data is then used to fine-tune the high-level policy, as detailed in~\mysecref{subsec:continuous_improvement}.
The specifics of the data, including the base dataset and the post-training dataset, are provided in Appendix~\ref{sec:appendix:data}.

\section{Experiments}
\label{sec:experiments}

Since \ours proposes a hierarchical system that interfaces the high-level and low-level policies through natural language, we want to answer the following questions: (1) Can human interventions through natural language change robot behavior meaningfully to improve task success on-the-fly? (2) Can these verbal corrections improve the autonomous performance of the robotic system on long-horizon tasks? (3) How does the hierarchical approach in \ours compare with non-hierarchical imitation learning methods? Further, we evaluate the importance of learning and improving a high-level policy in \ours, by ablating it with scripted policies and VLMs like GPT-4V.
\subsection{Robot Setup and Tasks}
\label{subsec:tasks}

\begin{figure*}[t]
    \centering
    \includegraphics[width=\textwidth]{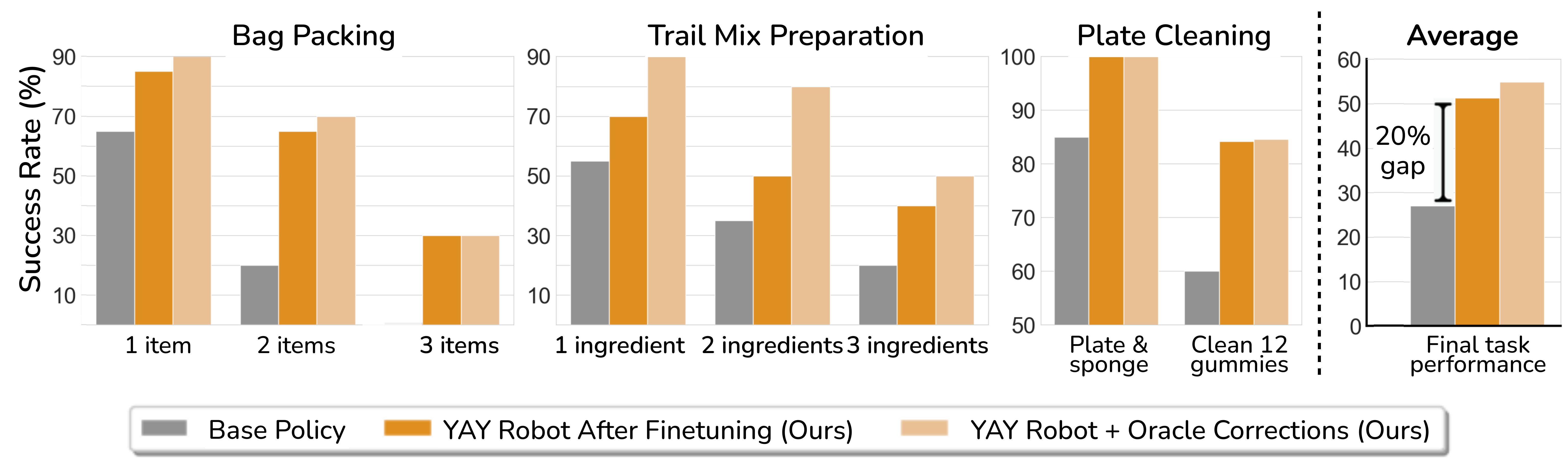}
    \vspace{-3mm}
    \caption{\textbf{Quantitative Evaluations.} Our system demonstrates a 20\% improvement in success rates over the base policy across three long-horizon bi-manual manipulation tasks, attributed to language corrections. These language corrections not only improve task success in real-time, but also substantially enhance the autonomous policy’s performance at each stage of the tasks through fine-tuning.
    }
    \label{fig:key_results}
    \vspace{-5mm}
\end{figure*}

We conduct experiments on a set of real-robot tasks using ALOHA, a low-cost, open-source bimanual hardware setup proposed in~\citet{zhao2023learning}. 
The setup includes a 14-dimensional action space corresponding to bimanual target joint positions and an observation space comprising RGB images from four cameras. 
Two cameras are mounted on the wrists for detailed object views during fine manipulation, and the other two are positioned at the front and top for broader perspectives. 
The demonstration data includes these camera feeds and the robots' joint positions at 50Hz.
Further hardware specifics can be found in prior works by ALOHA 2 Team~\citep{aloha2_2024} and~\citet{zhao2023learning}.

We select three long-horizon tasks emphasizing precision, coordination, and contact-rich manipulation, involving challenging manipulation of deformable and transparent objects~(\myfigref{fig:task}). 
Each task is designed to reflect useful, everyday robotic applications:

\noindent \textbf{Bag Packing:}
This task involves packing three randomly positioned objects -- a sharpie, a tape holder, and a sponge -- into a small ziploc bag. We selected these items specifically as they present distinct challenges: the sharpie and tape holder require high precision for successful grasping and insertion (e.g., the tape holder has a clearance of \~{}2mm), while the sponge, due to its size and deformable nature, may require squeezing to fit into the bag.
The robot must perform a series of skills: picking up each object, inserting it into a bag, and arranging items within the bag to avoid obstruction to subsequent objects.
Additionally, this task demands correction skills such as directional adjustments (right, left, towards, away, higher, lower), gripper rotation (clockwise, counter-clockwise), shaking the bag, and using the object to widen the bag opening, among others.

\noindent \textbf{Trail Mix Preparation:}
This task involves creating a bag of trail mix from various ingredients (almonds, peanuts, M\&M's, cranberries, and banana chips). 
Essential skills include using tools like a metal scoop, as well as precise maneuvers to scoop specific ingredients, adjust the quantity as per user instruction, and pour them into a transparent ziploc bag. 
Additionally, it involves corrections like adjusting the scoop's tilt and orientation, manipulating the bag for easier filling, and avoiding contact with food container edges.

\noindent \textbf{Plate Cleaning:} 
This task emulates a daily task in the real world, the cleaning of a dirty plate, requiring both precise and contact-rich manipulation and coordination from both arms for effective cleaning. For hygiene and safety reasons, we choose gummy bear candies that can stick to the plate surface as the food debris to be scrubbed. This task consists of four sets of skills: picking up the plate, reaching and grasping the sponge, moving the plate to a bowl, and wiping the gummy bears off the plate using the sponge. This task exhibits two common modes of failure: imperfect reaching or grasping of the plate and the sponge, and insufficient coverage of the wiping motion. Hence, correction skills such as ``pick up the sponge again", ``wipe the right side", and ``clean the bottom" are crucial to clean the plate thoroughly.

\subsection{Methods and Evaluation Protocol}
\label{subsec:baselines}

In our study, we first train the low-level and high-level policies (\textbf{Base Policy}).
Next, we consider \ours in the interactive setting, where the high-level policy controls the low-level policy through natural language instructions but a human can override the high-level command and issue a natural language instruction (\textbf{\ours + Oracle Corrections (Ours)}). This method allows us to evaluate the ability of the system to modulate its behavior on-the-fly and whether such interventions can improve the task performance, and we expect the task performance in this setting to be an upper bound on the performance of the best high-level policy.
The final step for \ours finetunes the high-level policy with data from human interventions collected in the shared autonomy setup, according to~\mysecref{sec:method}. We evaluate the autonomous performance of \ours after fine-tuning (\textbf{\ours after Finetuning (Ours)}), with no additional human interventions. 
Notably, for all evaluation tasks, operators without a background in computer science or robotics research collect the teleoperation data and provide verbal feedback to \ours, which also helps us assess the viability for real-world use in everyday settings by non-expert users.

To compare the performance of hierarchical policies with flat policies for long-horizon tasks, we benchmark our method against Action Chunking with Transformers (ACT~\citep{zhao2023learning}; \textbf{Flat-BC}), a state-of-the-art imitation learning method. We train ACT on the same training data as all the baselines above.

We conduct 20 trials for each evaluation. 
Given the long-horizon nature of the tasks, we also measure the sub-task success rate to provide more granular insights.
We detail the success criteria for each task in Appendix~\ref{subsec:appendix:success}.
Except for \textbf{\ours + Oracle Corrections}, which allows \emph{verbal} interventions from humans but no physical interventions, all evaluations are fully autonomous for both low-level and high-level policies.

\subsection{Key Results}
\label{subsec:results}

\textbf{Language corrections enhance task success on the fly.}
As shown in~\myfigref{fig:key_results}, our experiments show significant improvements in success rates at each stage when incorporating real-time language corrections: 25\%-50\% in Bag Packing, 30\%-45\% in Trail Mix Preparation, and 15\%-25\% in Plate Cleaning. 
Despite the high-level policy being trained on the entire dataset, including correction skills, we observe that it predominantly predicts task-specific instructions and rarely issues corrections prior to fine-tuning, as shown in our supplementary video.

We also demonstrate in our supplementary video how simple human language feedback helps address common failures -- for instance, in the Bag Packing task, when the robot struggles with precise grasping, a simple human direction like ``move to the left'' enables the robot to adjust its position for a successful grasp. 
Furthermore, language feedback is particularly useful in out-of-distribution (OOD) scenarios, such as when the sharpie slides beneath the transparent bag during insertion or when objects entangle with the bag opening due to its deformability.
In scenarios involving irreversible actions, such as in Trail Mix Preparation, humans can provide useful preemptive interventions, such as letting the robot adjust the angle of the end effector or the position of the bag to prevent pouring nuts on the table.
We save such interventions in the dataset to fine-tune the high-level policy, so that the robot can self-correct in subsequent interactions.

\begin{figure}[t]
\vspace{-2mm}
    \centering
    \vspace{-3mm}
    \includegraphics[width=0.8\linewidth]{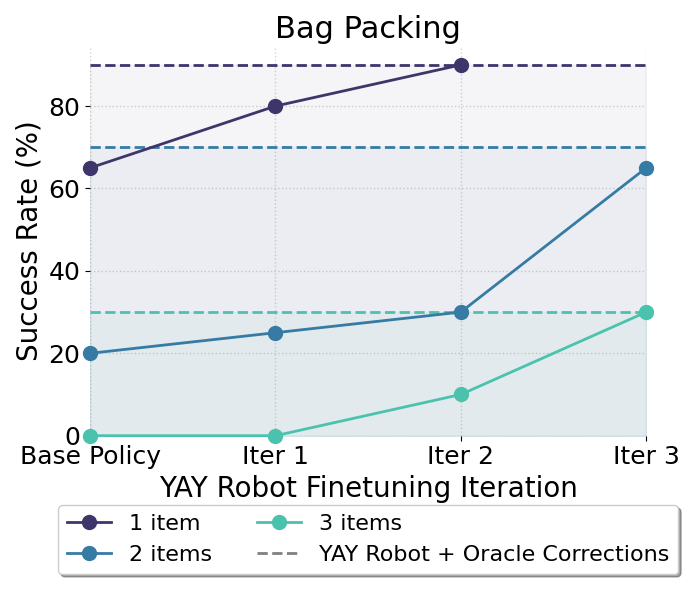}
    \vspace{-3mm}
    \caption{\textbf{Iterative Improvement.} \ours's success rates for packing different numbers of items show significant improvement with each iteration of user verbal feedback collection and fine-tuning, approaching the oracle's performance (dashed lines) at each stage of the task. }
    \label{fig:improvement}
    \vspace{-7mm}
\end{figure}

\begin{figure}[t]
\vspace{-3mm}
    \centering
    \includegraphics[width=0.95\linewidth]{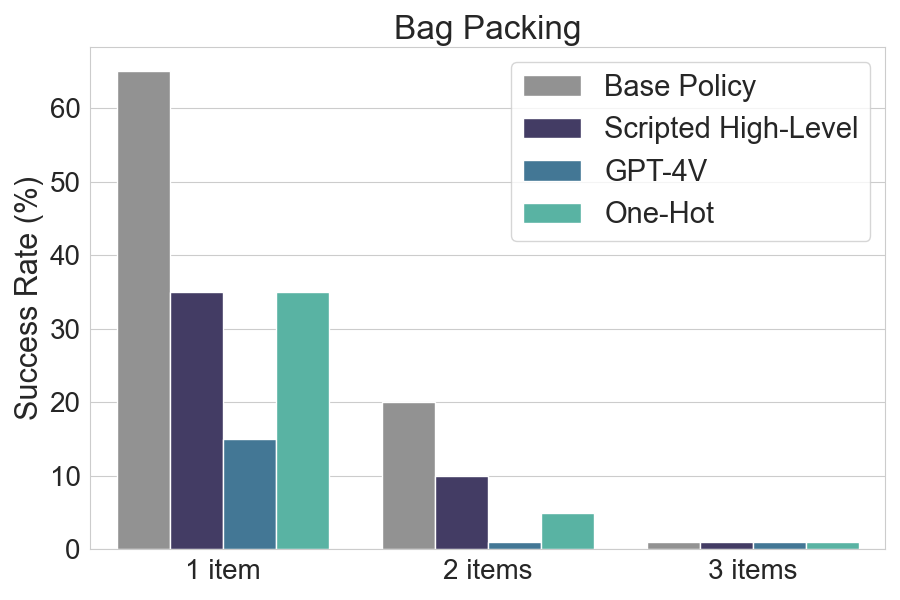}
    \vspace{-3mm}
    \caption{\textbf{Ablation on High-Level Policy.} Our results show that 1) replacing a learned high-level policy with a scripted one leads to performance decrease, 2) off-the-shelf VLM performs poorly on complex long-horizon task, and 3) replacing language with one-hot encodings hurts model performance.}
    \label{fig:ablation}
    \vspace{-3mm}
\end{figure}

\begin{table}[t]
\centering
\begin{tabular}{@{}c|c|c|c@{}}
\toprule
\textbf{Task} & \textbf{Substage} & \textbf{Base Policy} & \textbf{Flat BC} \\ 
\midrule
\multirow{3}{*}{Bag Packing} & 1 item & $\bestscore{65}$ & 25 \\
                             & 2 items & $\bestscore{20}$ & 0 \\
                             & 3 items & 0 & 0 \\
\midrule
\multirow{3}{*}{Trail Mix Preparation} & 1 ingredient & 55 & $\bestscore{60}$ \\
                                       & 2 ingredients & $\bestscore{35}$ & 20 \\
                                       & 3 ingredients & $\bestscore{20}$ & 5 \\
\midrule
\multirow{2}{*}{Plate Cleaning} & Prepare plate \& sponge & $\bestscore{85}$ & 80 \\
                                & Clean 12 gummies & $\bestscore{60}$ & 55 \\
\bottomrule
\end{tabular}
\caption{\textbf{Comparison to Flat Policy.} Overall, our hierarchical approach achieves higher success rates (\%) than the non-hierarchical imitation learning method on long-horizon tasks.}
\vspace{-8mm}
\label{table:flat_bc}
\end{table}

\begin{figure*}[t]
\vspace{-1em}
    \centering
    \includegraphics[width=\textwidth]{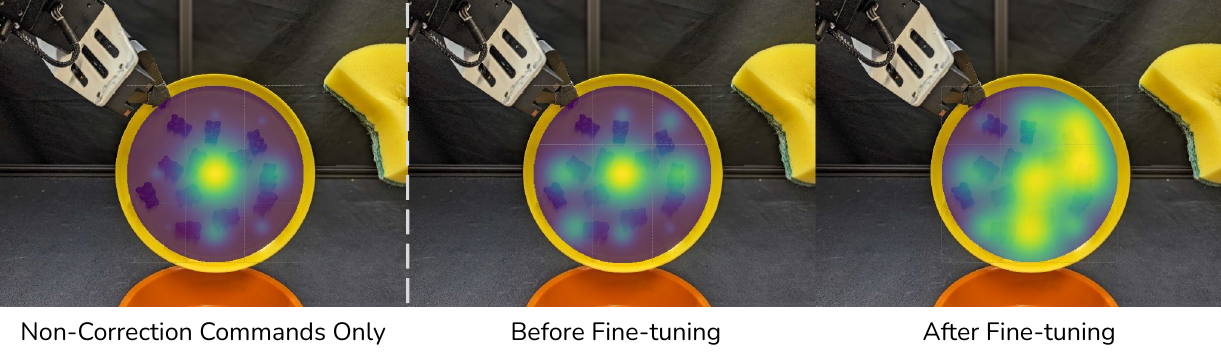}
    \vspace{-5mm}
    \caption{%
    \textbf{Policy Proficiency Increase through Fine-Tuning.} Through heatmaps, we visualize the cleaning efficacy across the plate surface, where brighter areas denote higher frequencies of effective wiping. \ours demonstrates wider cleaning coverage after fine-tuning the high-level policy with human verbal feedback.
    }
    \vspace{-4mm}
    \label{fig:plate_heatmaps}
\end{figure*}

\begin{figure}[t]
    \centering
    \includegraphics[width=\columnwidth]{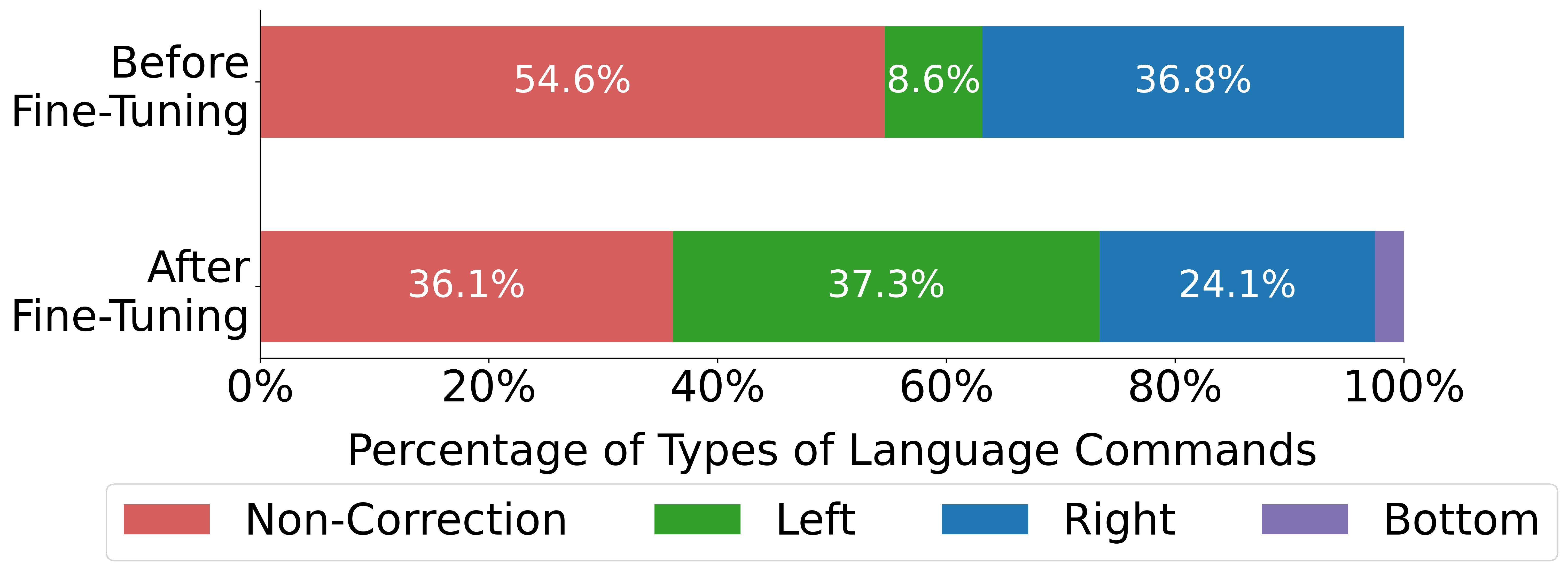}
    \vspace{-1em}
    \caption{
    \textbf{Non-Correction vs. Correction Skill Ratio.} Our analysis illustrates a shift from non-correction to corrective cleaning commands -- cleaning left, right, and bottom -- following policy fine-tuning, which leads to a notable enhancement in cleaning coverage (\myfigref{fig:plate_heatmaps}), and thus more effective cleaning.
    }
    \vspace{-8mm}
    \label{fig:plate_bars}
\end{figure}

\textbf{Finetuning on language corrections improves autonomous policy performance.}
Compared to the base policy, finetuning with our approach improves success rates at each stage by 20\%-45\% in Bag Packing, 15\%-20\% in Trail Mix Preparation, and 15\%-25\% in Plate Cleaning.
After fine-tuning, the high-level policy starts to autonomously generate corrections, a behavior that evolves through iterative post-training~(\myfigref{fig:improvement}).
For example, in the Bag Packing task, the high-level policy begins to instruct the robot to ``open the gripper wider'' before reattempting to grasp, ``rotate the right gripper clockwise'' for successful insertion, or to ``wiggle'' when an object becomes entangled with the bag.
In the Plate Cleaning task, removing all food debris covering the surface requires multiple wipes covering different regions of the plate. 
Language corrections like ``clean the" + ``right", ``left", ``top", and ``bottom" allow the robot to focus on specific areas of the plate and thus maximizing the cleaning results.
We present qualitative and quantitative analysis of the policy behavior prior to and following fine-tuning in \myfigref{fig:plate_heatmaps} and \myfigref{fig:plate_bars}.

\textbf{The hierarchical policy outperforms the flat policy on long-horizon tasks.}
As shown in~\mytabref{table:flat_bc}, success rates of hierarchical policies are generally higher than Flat-BC. The difference between hierarchical and flat policies is more pronounced in later stages of each task, indicating the hierarchical policy's better handling of compounding errors.

\subsection{Ablation Studies}
\label{subsec:ablation}

Next, we explore the necessity of a learned high-level policy. 
We compare our approach with both a predefined sequence of language instructions (\textbf{Scripted High-Level}) and a state-of-the-art vision-language model, GPT-4V (\citet{openai2023gpt4}; \textbf{GPT-4V}), as high-level policies. 
The GPT-4V policy is informed about the environment, task specifics, camera setups, and useful language instructions and corrections relevant to the task through a carefully crafted prompt~(Appendix~\ref{subsec:appendix:gpt4}). 
To examine the role of language conditioning on performance, we substitute language embeddings with one-hot skill encodings (\textbf{One-Hot}; Appendix~\ref{subsec:appendix:onehot}). 
Additionally, we investigate the trade-off between data quality and quantity by comparing ACT trained on filtered data that has removed suboptimal behaviors against training with the complete dataset (\textbf{All-Data}).

\textbf{Scripted High-Level Policy:} 
We replace our learned high-level policy with a predefined sequence of language instructions. 
As shown in~\myfigref{fig:ablation}, we observe a marked decrease in task performance, with the policy performing worse than the Base Policy by up to 30\%. 
Despite providing the most optimal skill sequence observed in policy rollouts, the scripted policy's inability to react to mistakes and adapt to unforeseen scenarios during deployment leads to failures. 
This finding underscores the necessity of a high-level policy that can dynamically address failures, especially given the imperfections of the low-level policy in handling all possible deployment scenarios.

\begin{table}[t]
\centering
\begin{tabular}{@{}c|c|c|c@{}}
\toprule
\textbf{Task} & \textbf{Substage} & \textbf{Filtered Data} & \textbf{All Data} \\ 
\midrule
\multirow{3}{*}{Bag Packing} & 1 item & $\bestscore{25}$ & $10$ \\
                             & 2 items & $0$ & $\bestscore{5}$ \\
                             & 3 items & $0$ & $0$ \\
\midrule
\multirow{3}{*}{Trail Mix Preparation} & 1 ingredient & $\bestscore{60}$ & $55$ \\
                                       & 2 ingredients & $\bestscore{20}$ & $\bestscore{20}$ \\
                                       & 3 ingredients & $\bestscore{5}$ & $0$ \\
\bottomrule
\end{tabular}
\caption{\textbf{Ablation on Data.} The policy's performance deteriorates when trained on larger datasets of mixed quality compared to smaller datasets of higher quality.}
\vspace{-8mm}
\label{table:ablation_data}
\end{table}

\textbf{GPT4-V as a High-Level Policy:} 
We also explore using GPT4-V as a high-level policy.
While it generates plausibly sounding reasoning steps, it frequently errs in understanding spatial relationships and the state of object manipulation, such as instructing the robot to ``lower the gripper'' when it has already touched the table, or to ``release the sponge'' when doing so would result in the sponge falling outside, as most of it is still outside the bag. 
These errors persist even when GPT4-V is provided with visual inputs from most convenient camera angles from our phones. 
This suggests that VLMs, without training on interaction data, are not sufficiently reliable for being high-level policies in robotic tasks.
Notably, our approach is complementary to pretrained VLMs. 
It would be interesting to explore our approach of high-level policy fine-tuning with a pretrained VLM, which we leave to future work.

\textbf{One-Hot vs. Language Embedding:} 
Replacing language embeddings with one-hot encodings in our model leads to inferior performance, occasionally resulting in unreasonable outputs. 
Given the large and diverse set of instructions in our dataset (e.g., 1,200 unique strings for the Bag task), we hypothesize language is crucial in leveraging the semantic similarities among instructions for model learning.

\textbf{Data Quality vs. Quantity:} 
Finally, we assess the importance of data quality in training. 
As shown in~\mytabref{table:ablation_data}, while training with the entire dataset (All Data) offers more data points, we observe that the training loss is less stable or slightly higher, especially on more challenging tasks. 
Besides, the robot occasionally exhibits suboptimal behavior even when the observation appears to be in-distribution.
This highlights the need for high-quality data or careful data filtering in training effective robotic policies.

\section{Conclusion and Limitations}
\label{sec:conclusion}

We presented a framework for enabling robots to respond to and improve using verbal corrections. On three long-horizon bi-manual manipulation tasks, our system reaches a 20\% higher success rate than the base policy, where the only additional supervision comes from verbal corrections. Despite promising results, our approach has a number of important limitations. To successfully handle language corrections both on-the-fly and in continuous improvement, our system critically relies on a performant low-level policy that can successfully react to many distinct language commands. Indeed, the performance of our approach with on-the-fly language corrections, which is essentially a near-optimal high-level policy, is far from perfect and not far above the performance using our fine-tuned high-level policy. This suggests that further performance improvements must come from improving the performance and flexibility of the low-level policy. Large-scale language-conditioned imitation learning approaches~\cite{jang2022bc,brohan2022rt1}, including methods that leverage vision-language models~\cite{brohan2023rt}, have shown promise in both improving performance and expanding the vocabulary of language-conditioned policies. Beyond verbal corrections, it may also be beneficial for people to communicate in other ways, e.g. by pointing~\cite{Constantin_2023_ICCV} or making other gestures. Our system is not well-equipped to handle such non-verbal communication. Looking forward, we hope that future research can further enable robots to improve with natural forms of human supervision, ultimately towards empowering anyone to help teach robots. %

\section*{Acknowledgments}
This work is supported by Sloan, Schmidt Futures, Boston Dynamics AI Institute, ONR grants N00014-20-1-2675, N00014-21-1-2685, and N00014-20-1-2383, NSF FRR IIS-2150826, and ARL DCIST CRA W911NF-17-2-0181. 
Tony Zhao is supported by Stanford Robotics Fellowship sponsored by FANUC.
We would like to thank Johnny Liang and Ethan Foster for data collection and policy evaluation. We especially thank Sidd Karamcheti for constructive feedback on the draft. We also thank Chen Wang, Moo Jin Kim, Eric Mitchell, Jonathan Yang, and Annie Chen for thought-provoking discussions.

\bibliographystyle{plainnat}

\newpage
\section*{Appendix}
\subsection{Data}
\label{sec:appendix:data}

\subsubsection{Base Dataset}
For each of the three tasks, we collect a base dataset consisting of human teleoperated trajectories, annotated by language. During data collection, the teleoperator presses down onto a USB foot paddle and records the audio for verbal commands, such as ``pick up the sponge'', using a USB microphone. Then, the teleoperator releases the foot paddle and controls the leader arms to provide the corresponding bimanual demonstrations to the puppet arms, as in ALOHA~\citep{zhao2023learning}. In \mytabref{table:base_data_summary}, we detail the number of trajectories and language commands collected for each task during this stage of data collection.

\begin{table}[h]
\centering
\resizebox{\columnwidth}{!}{
\begin{tabular}{@{}lccc@{}}
\toprule
& \textbf{Bag Backing} & \textbf{Trail Mix} & \textbf{Plate Cleaning} \\ 
\midrule
\textbf{Trajectories} & 1170 & 317 & 265 \\
\textbf{Traj. Length (timesteps)} & 2000 - 5000 & 3000 & 1200 \\
\textbf{Skill Segments} & 41517 & 7008 & 3236 \\
\textbf{Unique Commands} & 1054 & 104 & 33 \\
\textbf{Commands Appeared $\geq$ 3 Times} & 862 & 47 & 25 \\
\bottomrule
\end{tabular}
}
\caption{\textbf{Base Dataset.} Summary of the number of trajectories, trajectory length, the number of language-annotated skill segments, and language command metrics across different tasks. ``Unique Commands'' denote the number of unique language strings in the dataset.} 
\label{table:base_data_summary}
\end{table}

\subsubsection{Post-Training Dataset}
For high-level policy fine-tuning, we collect language-only human intervention data using the USB microphone device mentioned above. 
In \mytabref{table:pt_data_summary}, we describe the aggregated dataset gathered after 2-3 iterations of post-training.
The post-training dataset has significantly fewer skill segments compared to the base dataset.

\begin{table}[h]
\centering
\resizebox{\columnwidth}{!}{
\begin{tabular}{@{}lccc@{}}
\toprule
& \textbf{Bag Backing} & \textbf{Trail Mix} & \textbf{Plate Cleaning} \\ 
\midrule
\textbf{Iterations} & 3 & 3 & 2 \\
\textbf{Skill Segments} & 2028 & 292 & 348 \\
\textbf{Ratio to the Bast Dataset} & 0.048 & 0.042 & 0.108 \\
\textbf{Unique Commands} & 120 & 36 & 27 \\
\textbf{Commands Appeared $\geq$ 3 Times} & 92 & 17 & 18 \\
\bottomrule
\end{tabular}
}
\caption{\textbf{Post-Training Dataset.} Summary of the number of skill segments and language commands in the post-training dataset for each task. The post-training dataset is significantly smaller than the base dataset -- the number of skill segments is 4\%-11\% of those in the base dataset.}
\label{table:pt_data_summary}
\end{table}

\subsubsection{Data Visualization}
\begin{figure*}[t]
    \centering
    \includegraphics[width=\textwidth]{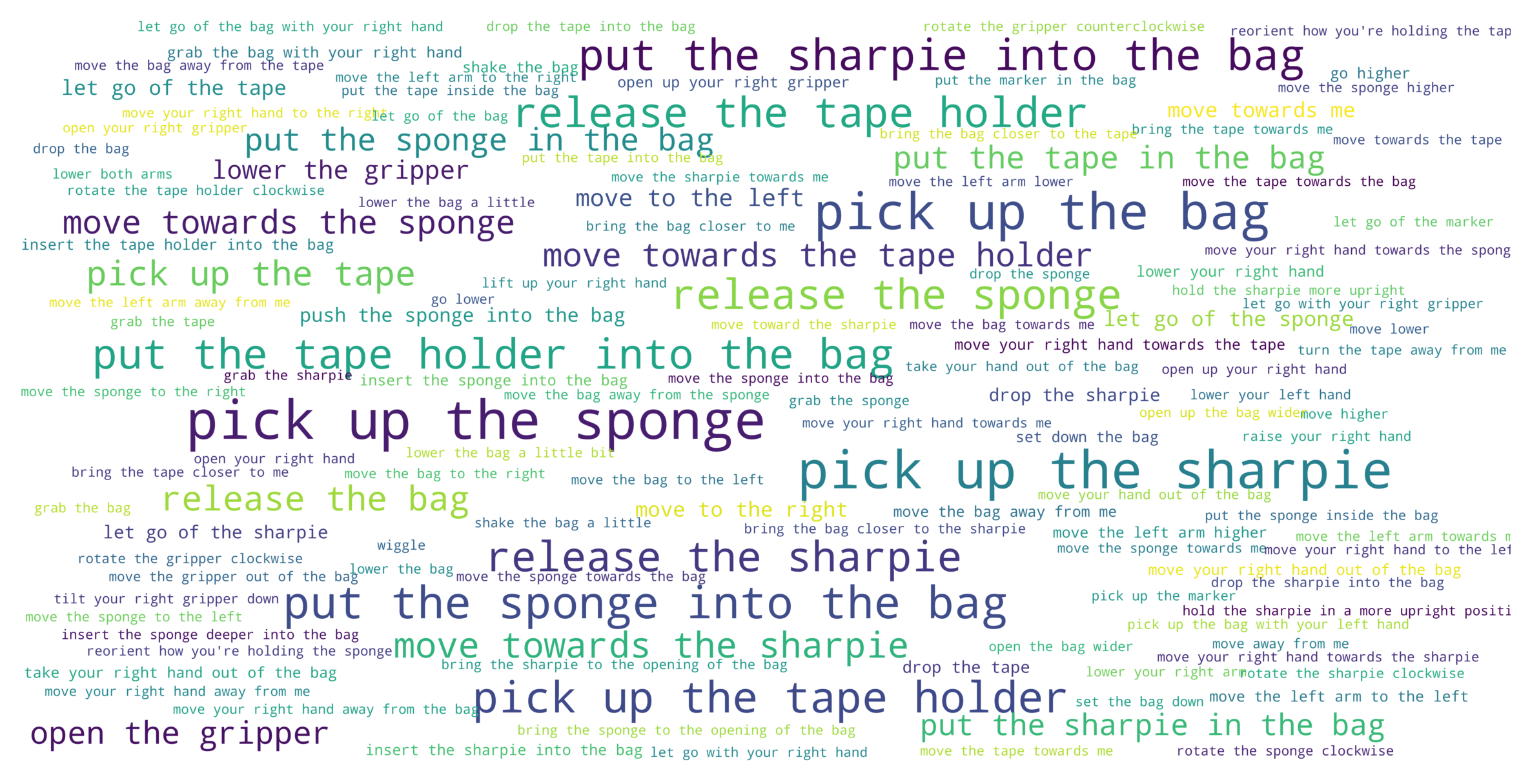}
    \vspace{-3mm}
    \caption{Word cloud of the most frequent 200 commands in bag packing datasets.}
    \label{fig:word_cloud}
\end{figure*}

We visualize the language commands in the bag packing task by making a word cloud out of the most frequent 200 commands in the datasets, as shown in~\myfigref{fig:word_cloud}.
The most frequently-appeared skills are task-oriented. Each correction skill has a lower frequency, but the correction skills are more diverse (e.g. ``wiggle'', ``rotate the gripper clockwise'', ``shake the bag a little'').

\subsection{Experiment Details}
\label{sec:appendix:experiment}

\subsubsection{Task Success Criterion}
\label{subsec:appendix:success}

We evaluate each method for 20 trials for all three tasks and report their success rates. The success criteria for each task are defined as follows:
\begin{enumerate}
    \item Bag Backing:
        \begin{itemize}
            \item ``1 item'' is successful if the robot successfully picks up the bag, grasps one of the objects, and inserts it into the bag.
            \item ``2 items'' is successful if the robot successfully picks up the second object and inserts it into the bag without making the first item fall out of the bag.
            \item ``3 items'' is successful if the robot successfully picks up the third object, inserts it into the bag without making the first or the second item fall out of the bag, and releases the bag without making any object fall out of the bag.
        \end{itemize}
    \item Trail Mix Preparation
        \begin{itemize}
            \item ``1 ingredient'' is successful if the robot successfully picks up the bag, brings the bag closer to the user, picks up the scoop, scoops one ingredient, and pours it into the bag.
            \item ``2 ingredients'' is successful if the robot successfully moves the gripper outside the bag, scoops the second ingredient, and pours it into the bag. If there exists any spill, the spill should be less than 1/4 of the total amount within the scoop.
            \item ``3 ingredients'' is successful if the robot successfully moves the gripper outside the bag, scoops the third ingredient, pours it into the bag, and releases the bag.
        \end{itemize}
    \item Plate Cleaning:
        \begin{itemize}
            \item Preparing the plate is successful if the robot grasps the plate and lifts up the plate mid-air. 
            \item Preparing the sponge is successful if the robot grasps the sponge and lifts up the sponge mid-air.
            \item Preparing for cleaning is successful if the robot holds and moves the plate to the bowl, and moves the sponge near the plate.
            \item Cleaning the plate success rate is measured by how many gummies are cleaned out of a total of 12 gummies on the plate.
        \end{itemize}
\end{enumerate}

\subsubsection{One-Hot}
\label{subsec:appendix:onehot}

We index all the 1054 unique language commands in the bag packing dataset. 
The model predicts logits for the indices (i.e., the output of the high-level policy is now a 1054-dimensional vector instead of language embeddings).
We then train the model with cross-entropy loss.
At the inference time, we select the index that has the highest probability (argmax).

\subsubsection{GPT-4V}
\label{subsec:appendix:gpt4}

To test the capability of pre-trained VLMs in reasoning about complex, long-horizon tasks, we prompt GPT-4V to provide instruction for the language-conditioned low-level policy. 
We use the following prompt in \myfigref{fig:gpt4_prompt} to inform GPT-4V about the task, camera set-up, and the most useful instructions and corrections for the bagging task.

\begin{figure*}[t]
    \centering
    \includegraphics[width=\textwidth]{figures/gpt4_prompt.pdf}
    \vspace{-3mm}
    \caption{Prompt to GPT-4V for the multi-camera setup. It includes the task information, camera setup, and 50 most useful instructions for the task.
    }
    \label{fig:gpt4_prompt}
    \vspace{-5mm}
\end{figure*}

\begin{figure*}[h]
\vspace{-2mm}
    \centering
    \vspace{-3mm}
    \includegraphics[width=0.7\textwidth]{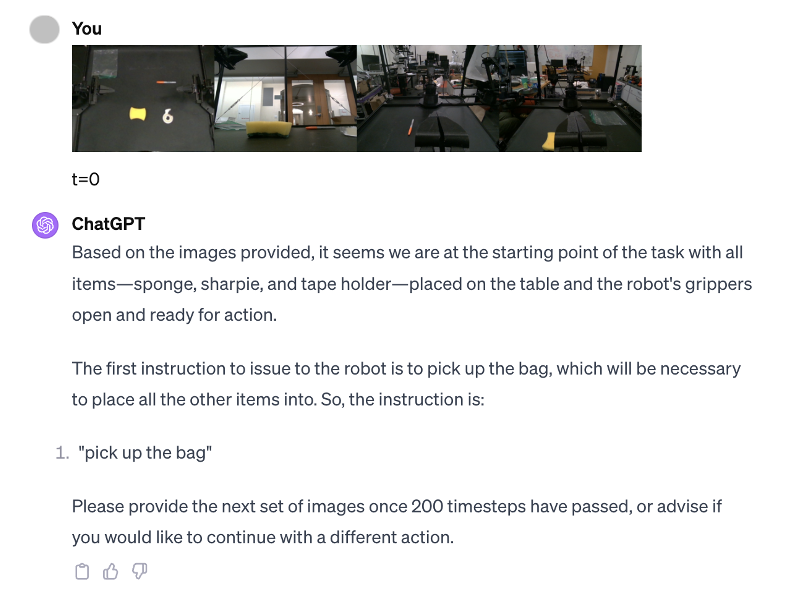}
    \vspace{-3mm}
    \caption{GPT-4V makes a reasonable choice at the beginning of the task. }
    \label{fig:gpt_start}
    \vspace{-7mm}
\end{figure*}

\begin{figure*}[h]
\vspace{-2mm}
    \centering
    \vspace{-3mm}
    \includegraphics[width=0.7\textwidth]{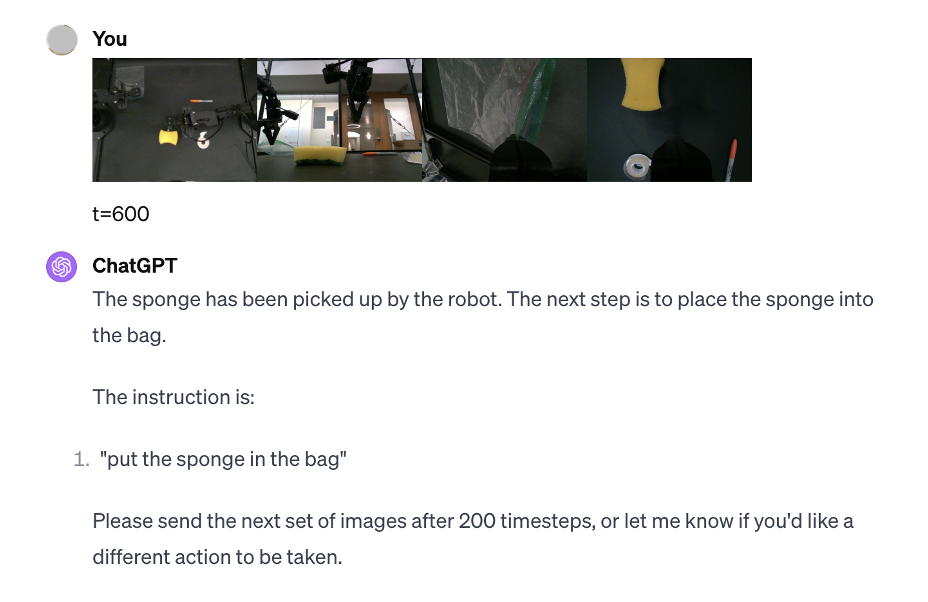}
    \vspace{-3mm}
    \caption{As the robot proceeds, the spatial reasoning of the VLM becomes incorrect. }
    \label{fig:gpt_off}
    \vspace{-7mm}
\end{figure*}

\begin{figure}[H]
    \centering
    \includegraphics[width=0.83\linewidth]{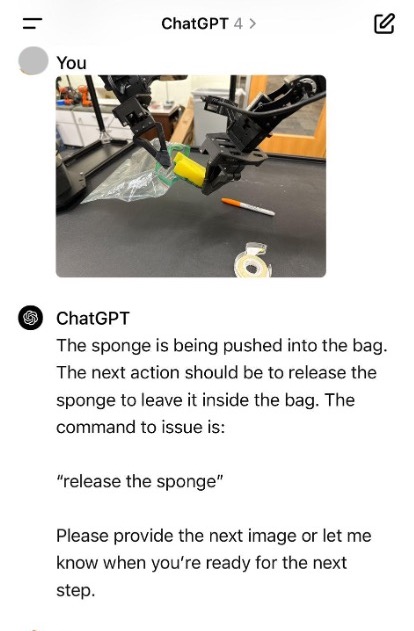}
    \vspace{-3mm}
    \caption{GPT-4V is limited by the lack of knowledge regarding specific robotic capabilities. }
    \label{fig:gpt_cam}
    \vspace{-7mm}
\end{figure}

GPT-4V is able to make reasonable predictions at the beginning, as shown in \myfigref{fig:gpt_start}.
However, usually after 2-3 reasonable predictions, its spatial reasoning becomes incorrect. 
For example, in \myfigref{fig:gpt_off}, even though the sponge is still on the table and far from the robot gripper, the model reasons that ``The sponge has been picked up by the robot'' and asks the robot to ``put the sponge in the bag''.
To check whether the suboptimal behavior is caused by the confusion over multiple camera observations, we tried to feed GPT-4V with only the third-person view from the front camera and adjust the prompt accordingly. 
It turns out that the model makes similar mistakes.
\begin{figure*}[h]
\vspace{-2mm}
    \centering
    \vspace{-3mm}
    \includegraphics[width=0.7\textwidth]{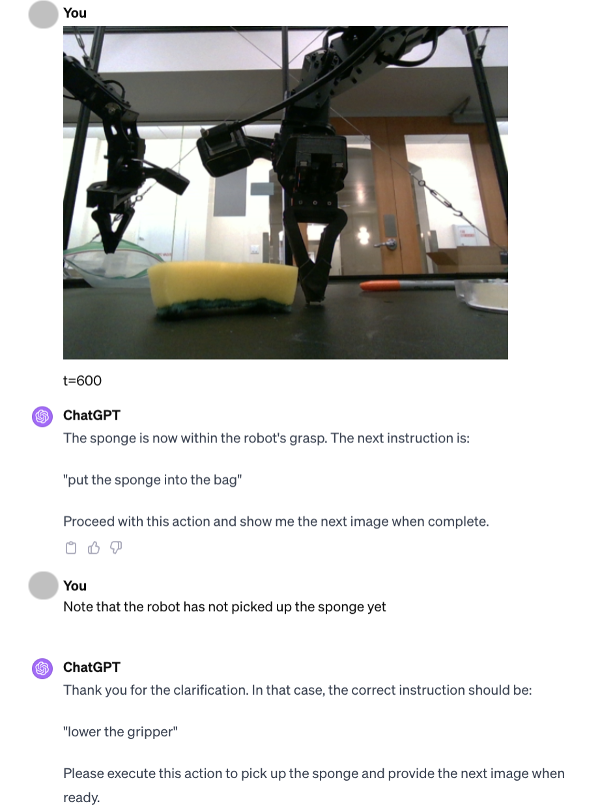}
    \caption{In the single-camera setup, the model still makes incorrect reasoning about the spatial relationship among the sponge, gripper, and table.}
    \label{fig:gpt_front}
\end{figure*}
As shown in \myfigref{fig:gpt_front}, the model similarly predicts ``put the sponge into the bag'' when the robot has not picked up the sponge yet. 
When we remind it of this fact through text, GPT-4V changes its prediction to ``lower the gripper'', while the gripper is already touching the table and cannot get lower.
Additionally, we tried to make spatial reasoning easier by providing image observations from the most convenient camera angles (taken by phone). 
However, the model likewise makes mistakes in understanding spatial relationships, such as taking an object that is still outside the bag to be inside the bag 
(\myfigref{fig:gpt_cam}).
Given that changing camera angles did not help, we used the same multi-camera set-up as our learned high-level policy for evaluation.

\end{document}